\PassOptionsToPackage{colorlinks=true,linkcolor = blue,urlcolor=blue}{hyperref}
\documentclass[sigconf]{acmart}

\usepackage{microtype}
\usepackage{graphicx}
\usepackage{subfigure}
\usepackage{booktabs} \usepackage{tabularx} \usepackage{amsfonts}
\usepackage{xcolor}
\usepackage{xspace}
\usepackage{enumitem}
\usepackage{algorithm}
\usepackage{algorithmic}

\definecolor{cycle1}{RGB}{235,172,35}
\definecolor{cycle2}{RGB}{184,0,88}
\definecolor{cycle3}{RGB}{0,140,249}
\definecolor{cycle4}{RGB}{0,110,0}
\definecolor{cycle5}{RGB}{0,187,173}
\definecolor{cycle6}{RGB}{209,99,230}
\definecolor{cycle7}{RGB}{178,69,2}
\definecolor{cycle8}{RGB}{255,146,135}
\definecolor{cycle9}{RGB}{89,84,214}
\definecolor{cycle10}{RGB}{0,198,248}
\definecolor{cycle11}{RGB}{135,133,0}
\definecolor{cycle12}{RGB}{0,167,108}
\definecolor{cyclegray}{RGB}{189,189,189}

\usepackage{tikz}
\usetikzlibrary{fit,positioning,calc,shapes,backgrounds,shapes.geometric,arrows.meta,patterns}
\usepackage{pgfplots}
\pgfplotsset{compat=1.16}
\usepgfplotslibrary{fillbetween, groupplots}

\usepackage{listings}
\newcommand{\friendster}{Friendster\xspace}
\newcommand{\papers}{OGBN-Papers100m\xspace}
\newcommand{\dwtpu}{HUGE-TPU\xspace}
\newcommand{\dwcpu}{HUGE-CPU\xspace}
\newcommand{\bg}{PyTorch-BigGraph\xspace}
\newcommand{\lne}{LightNE\xspace}
\newcommand{\ie}{InstantEmbedding}
\newcommand{\snr}{{SNR}\xspace}
\newcommand{\replicabatch}{\texttt{per\_replica\_batch\_size}\xspace}
\newcommand{\nwalks}{\texttt{num\_walks\_per\_node}\xspace}
\newcommand{\walklen}{\texttt{walk\_length}\xspace}
\newcommand{\sourceid}{\texttt{source\_id}\xspace}
\newcommand{\destid}{\texttt{destination\_id}\xspace}
\newcommand{\cocounts}{\texttt{co\_counts}\xspace}
\newcommand{\nnpp}{\texttt{num\_neg\_per\_pos}\xspace}
\newcommand{\sbmhm}{SBM-100M}
\newcommand{\sbmb}{SBM-1000M}

\AtBeginDocument{}

\copyrightyear{2023}
\acmYear{2023}
\setcopyright{rightsretained}
\acmConference[KDD '23]{Proceedings of the 29th ACM SIGKDD Conference on Knowledge Discovery and Data Mining}{August 6--10, 2023}{Long Beach, CA, USA}
\acmBooktitle{Proceedings of the 29th ACM SIGKDD Conference on Knowledge Discovery and Data Mining (KDD '23), August 6--10, 2023, Long Beach, CA, USA}
\acmDOI{10.1145/3580305.3599840}
\acmISBN{979-8-4007-0103-0/23/08}

\makeatletter
\gdef\@copyrightpermission{
  \begin{minipage}{0.3\columnwidth}
   \href{https://creativecommons.org/licenses/by/4.0/}{\includegraphics[width=0.90\textwidth]{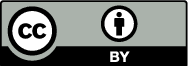}}
  \end{minipage}\hfill
  \begin{minipage}{0.7\columnwidth}
   \href{https://creativecommons.org/licenses/by/4.0/}{This work is licensed under a Creative Commons Attribution International 4.0 License.}
  \end{minipage}
  \vspace{5pt}
}
\makeatother


\begin{document}

\title{HUGE: Huge Unsupervised Graph Embeddings with TPUs}

\author{Brandon A. Mayer}
\email{bmayer@google.com}
\affiliation{\institution{Google Research}
  \country{USA}}
\author{Anton Tsitsulin}
\email{tsitsulin@google.com}
\affiliation{\institution{Google Research}
  \country{USA}}
\author{Hendrik Fichtenberger}
\email{fichtenberger@google.com}
\affiliation{\institution{Google Research}
  \country{Switzerland}}
\author{Jonathan Halcrow}
\email{halcrow@google.com}
\affiliation{\institution{Google Research}
  \country{USA}}
\author{Bryan Perozzi}
\email{bperozzi@acm.org}
\affiliation{\institution{Google Research}
  \country{USA}}

\begin{abstract} Graphs are a representation of structured data that captures the relationships between sets of objects.
With the ubiquity of available network data, there is increasing industrial and academic need to quickly analyze graphs with billions of nodes and trillions of edges.
A common first step for network understanding is Graph Embedding, the process of creating a continuous representation of nodes in a graph.
A continuous representation is often more amenable, especially at scale, for solving downstream machine learning tasks such as classification, link prediction, and clustering.
A high-performance graph embedding architecture leveraging Tensor Processing Units (TPUs) with configurable amounts of high-bandwidth memory is presented that simplifies the graph embedding problem and can scale to graphs with billions of nodes and trillions of edges. We verify the embedding space quality on real and synthetic large-scale datasets.
\end{abstract}

 \maketitle
\section{Introduction}
Graph data naturally arises in many domains, including social, biological, and computer networks and the structure of the Web, and user-content interactions including purchase and content networks.
Graph can greatly vary in size -- in industrial applications, they often times grow to billions of nodes and trillions of edges in size. Making intelligent automated decisions with such large scale graphical data sets is extremely compute and storage intensive, making these tasks hard or impossible to solve using commodity hardware.

\begin{figure}
    \centering
    \includegraphics[width=.6\linewidth]{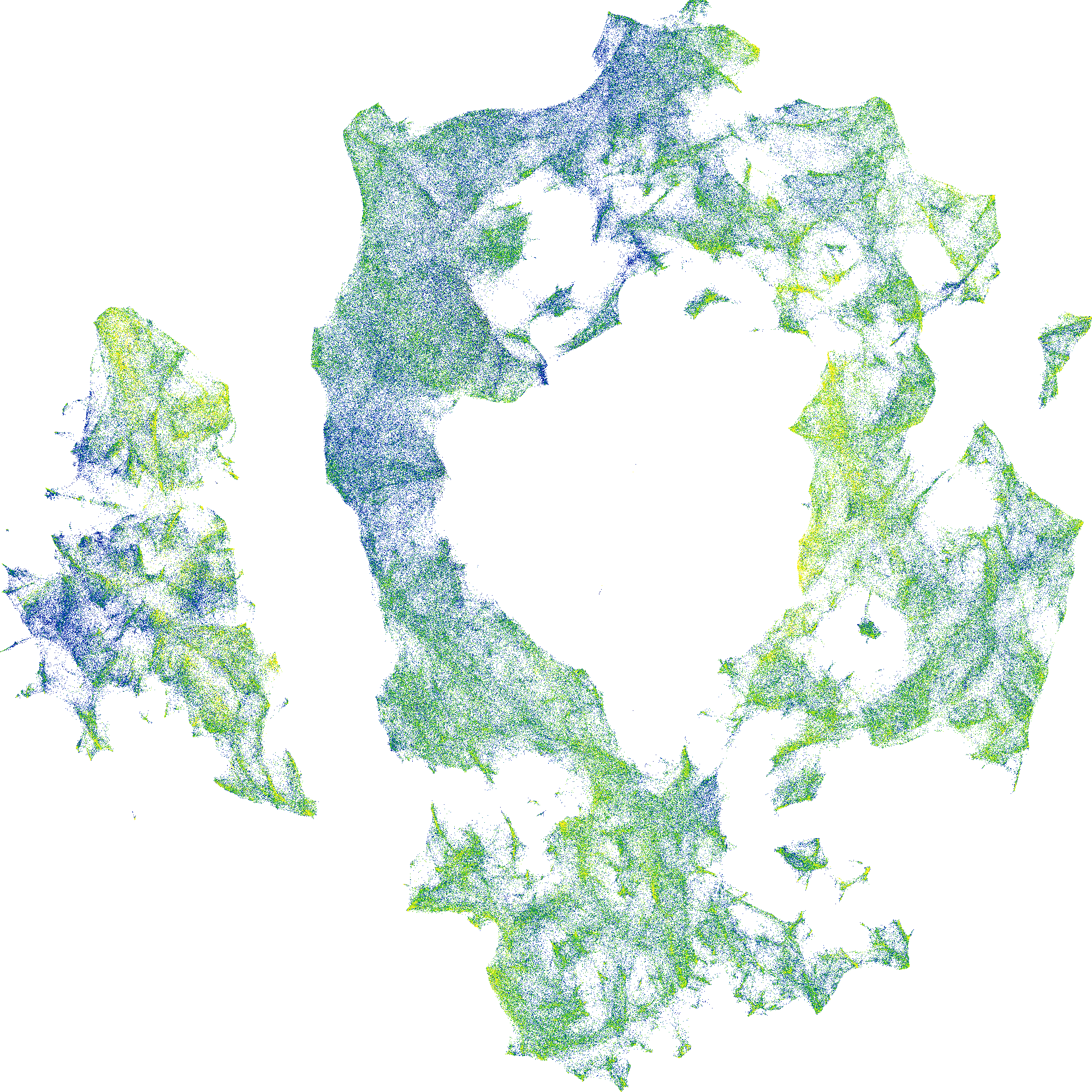}
    \vspace{-1em}
    \caption{HUGE can learn representations on extremely large graphs (billions of nodes) at Google. (Shown here: t-SNE projection of HUGE-TPU's Papers100M embeddings)\label{fig:viz}}
\vspace{-2em}
\end{figure}

Graph embeddings\footnote{Also referred to as \emph{node embeddings} in the literature.} are a common first step in graph understanding pipelines where every node in the graph is \emph{embedded}\cite{chen2018tutorial} into a common low-dimensional space.
These embeddings are then used for graph learning tasks such as node classification, graph clustering, and link prediction \cite{ chen2018tutorial, chami2022machine}  that can be solved with standard machine learning algorithms applied in the graph embedding space without having to develop specific algorithms to directly exploit the graph structure.
For example, approximate nearest neighbor search systems~\cite{avq_2020} can serve product recommendations using embeddings of user-item graphs.

Numerous methods, which we briefly review in Section~\ref{sec:relatedwork}, have been proposed in the literature.
One of the most popular methods, DeepWalk~\cite{perozzi2014deepwalk}, proposes to embed nodes via a shallow neural network trained to discriminate samples generated from random walks from random negatives.
Unfortunately, the process is memory-bound, as it requires random accesses for both random walk generation and updating the embedding table. 
At first glance, it does not scale beyond graphs larger than a couple of millions of nodes even in the distributed setting.

Or does it?
Distributed data processing workflows (such as Flume~\cite{flumejava} or \href{https://beam.apache.org/}{Apache Beam}) can be used to execute sampling strategies even for graphs that are too large to represent in memory on a single machine.
At the same time, specialized hardware such as Tensor Processing Units (TPUs), a custom ASIC introduced in~\citet{tpuWhitepaper}, have large amounts of high-bandwidth memory that enables high-throughput gradient updates.
In this work, we present a simple, TPU-based architecture for graph embedding that exploits the computational advantages of TPUs to embed billion-scale graphs\footnote{Open source implementation available at: \url{https://github.com/google-research/google-research/tree/master/graph_embedding/huge}}.
This architecture eliminates the need to develop complex algorithms to partition or synchronize the embedding table across multiple machines.

\begin{table*}[!t]
\centering
\caption{An overview of different approaches to scaling graph embeddings up. In this work, we demonstrate a system that works \emph{without} any of the yet mentioned techniques scaled to largest graphs. Scalability is given as an approximate graph size that can be processed by best-performing algorithm\&system combination in a day.}\label{tbl:related-work}
\newcolumntype{R}{>{\raggedleft\arraybackslash}X}
\newcolumntype{C}{>{\centering\arraybackslash}X}
\newcolumntype{L}{>{\hsize=.4\hsize}C}
\begin{tabularx}{\linewidth}{@{}LLCL@{}}
\toprule
Method Family & Speedup Technique & Reference Methods & Scalability \\
\midrule
neural & --- & DeepWalk, LINE, VERSE& 10M \\
neural & graph coarsening & HARP, MILE & 10M \\
neural & partitioning & BigGraph, EDGES, DeLNE & 1000M \\
\midrule
factorization & --- & NetMF, GraRep & 10k \\
factorization & matrix sparsification & NetSMF, NetMFSC, STRAP, NRP & 100M \\
factorization & implicit solvers & HOPE, AROPE & 100M \\
factorization & spectral propagation & ProNE, LightNE & 1000M \\
factorization & matrix sketching & FastRP, RandNE, NodeSketch, InstantEmbedding & 1000M \\
\bottomrule
\end{tabularx}
\end{table*}
 
More specifically, we propose a two-phase architecture. First random walks are generated and summarized via a distributed data processing pipeline. After sampling, graph embedding is posed as a machine learning problem in the style of DeepWalk~\cite{perozzi2014deepwalk}. We propose an unsupervised method for measuring embedding space quality, comparing the embedding space result to the structure of the original graph and show that the proposed system is competitive in speed and quality compared to modern CPU-based systems while achieving HUGE scale.

\noindent\textbf{Graph Embeddings at Google.}
Graph-based machine learning is increasingly popular at Google (\citet{graphmining2020}).
There are dozens of distinct model applications using different forms of implicit and explicit graph embedding techniques in many popular Google products.
Over time, these models have evolved from single-machine in-memory algorithms (similar to those dominating in academic literature) to more scalable approaches based on distributed compute platforms.
In this work we detail a relatively new and very promising extension of classic graph embedding methods that we have developed in response to the need for differentiable graph embedding methods which can operate with data at extreme scale.  \section{Background}\label{sec:background}
In this section, we first briefly review the related work in Section~\ref{sec:relatedwork}. We review DeepWalk~\cite{perozzi2014deepwalk}, which we use as a base for our high-performance embedding systems, in Section~\ref{sec:deepwalk}. We then proceed with describing two architectures for scaling Deepwalk graph embedding using commodity (\dwcpu{}) and TPU (\dwtpu{}) hardware that allow us to scale to huge graphs in Section~\ref{sec:tpus}.

\subsection{Related Work}\label{sec:relatedwork}

We now proceed to review the two basic approaches to embedding large graphs.
Over the past years, tremendous amount of work introduced various embedding methods as well as a myriad of techniques and hardware architectures to scale them up.
We summarize the related work in terms of the embedding approach, speedup techniques, and its expected scalability in Table~\ref{tbl:related-work}. 
\subsubsection{Graph Embedding Approaches}

We categorize embedding methods as either \emph{neural network}-based or \emph{matrix factorization}-based.
Regardless of the approach, each method employs, sometimes implicitly, a similarity function that relates each node to other nodes in the graph.
The best-performing methods depart from just using the adjacency information in the graph to some notion of random walk-based similarity, for example, personalized PageRank (PPR)~\cite{brin1998anatomy}.
A key insight for accelerating the computation of these similarities is that they are highly localized in the graph~\cite{andersen2006local}, meaning 2--3 propagation steps are enough to approximate them.

Neural embedding methods view node embeddings as parameters of a shallow neural network.
Neural methods optimize these parameters with stochastic gradient descent for either adjacency~\cite{tang2015line}, random walk~\cite{perozzi2014deepwalk}, or personalized PageRank~\cite{tsitsulin2018verse} similarity functions.
This optimization is done via sampling, and the updates to the embedding table are usually very sparse.
Thus, random memory access typically bounds the performance of these methods.

An alternative to gradient-based methods is to directly factorize the similarity matrix.
There are deep connections between neural and matrix factorization approaches~\cite{qiu2018network,tsitsulin2021frede}---essentially, for many node similarities the optimal solutions for neural and factorization-based embeddings coincide.
The main challenge to matrix-based methods is maintaining sparsity of intermediate representations.
For large graphs, one can not afford to increase the density of the adjacency matrix nor keep too many intermediate projections.

\subsubsection{Scaling Graph Embedding Systems}

There are several directions for speeding up embedding algorithms---some are tailored to particular methods while some are more general.
We now briefly review the most general speedup techniques.
Graph coarsening~\cite{chen2018harp,liang2021mile} iteratively contracts the graph, learns the embeddings for the most compressed level, and deterministically propagates the embeddings across the contraction hierarchy.
Graph partitioning methods~\cite{pbg,imran2020decentralized} distribute the computation across machines while attempting to minimize communication across machines.

Early approaches to matrix factorization~\cite{qiu2019netsmf,yin2019scalable,yang2019homogeneous} attempt to sparsify the random walk or PPR matrices.
Unfortunately, higher-order similarity matrices are still too dense for these embeddings methods to scale to very large graphs.
Leveraging specialized sparse numerical linear algebra techniques~\cite{saad2003iterative} proved to be a more fruitful approach.
Implicit solvers~\cite{ou2016asymmetric,zhang2018arbitrary} can factorize the matrix without explicitly materializing it in memory.
These methods are constrained to perform linear decomposition, which is not able to successfully account for structure of graphs.

Two families of techniques that produce most scalable embedding methods are spectral propagation and matrix sketching~\cite{woodruff2014sketching}.
Spectral propagation methods~\cite{zhang2019prone,qiu2021lightne} first compute some truncated eigendecomposition of the adjacency or the Laplacian matrix of a graph and then use these eigenvectors to simulate the diffusion of information.
Matrix sketching approaches approximate the similarity matrix, either iteratively~\cite{zhang2018billion,chen2019fast,yang2019nodesketch} or in a single pass~\cite{postuavaru2020instantembedding}.
The latter option is more scalable.

\subsubsection{Hardware-based Embedding Acceleration}

Compared to algorithmic advances, hardware-based acceleration has arguably received less attention.
\citet{zhu2019graphvite} proposes a hybrid system that uses CPU for sampling and GPU for training the embeddings.
Since the most RAM a single GPU can offer is in the order of 100 gigabytes, one can only train embeddings of 100 million node graphs on such systems.
\citet{wei2020distributed,yang2020edges} address this problem with partitioning to include more GPUs.
This approach requires tens of GPUs for a billion-node graph, which is prohibitive compared to scalable CPU-based systems, which can embed a billion-node graph on a single high-memory machine in hours.

Efficient computation of higher-order similarity is one aspect where hardware acceleration is currently lacking.
\citet{yang2021random,wang2020graphwalker} propose efficient systems for random walk generation for general hardware architectures.
However, in an absence of a suitable embedding method, these systems are not useful for graph embedding.

\subsection{DeepWalk}\label{sec:deepwalk}

Before describing our TPU embedding system, it is necessary to review DeepWalk~\cite{perozzi2014deepwalk}, which is the basic method for neural graph embedding.
DeepWalk adapts word2vec~\cite{mikolov2013distributed}, a widely successful model for embedding words, to graph data.
DeepWalk generates a ``corpus'' of short random walks; the objective of DeepWalk is to maximize the posterior probability of observing a neighboring vertex in a random walk within some specific window size.
To maximize this probability efficiently, it uses hierarchical softmax~\cite{morin2005hierarchical}, which constructs a Huffman tree of nodes based on their frequency of appearance, or a more computationally efficient approximation, negative sampling~\cite{gutmann2010noise}.
For each node that was observed within the window size from some node, DeepWalk picks $k\ll n$ uniformly at random as contrastive negative examples.

There are several computational problems with DeepWalk's architecture, which are to be solved if we are to scale DeepWalk to graphs with billions of nodes:
\begin{itemize}
    \item Random walk generation for large graphs is computationally prohibitive due to random memory accesses on each random walk step.
    \item Random walk corpus grows in size rapidly, growing much larger in size than the original sparse graph.
    \item Negative sampling-based optimization is also computationally prohibitive due to random memory accesses. If batched, each gradient update is bound to update a significant part of the embedding table.
\end{itemize}

To overcome the difficulties with random walk sampling, we present a distributed random walk algorithm in section~\ref{sec:sampling} that is routinely used at Google to scale random walk simulations to web-scale graphs.

\subsection{Tensor Processing Units}\label{sec:tpus}

We proceed with briefly reviewing TPU architecture highlighting the aspects critical for our graph embedding system.
A detailed review can be found in~\cite{tpuWhitepaper,jouppi21,jouppi23}.
TPUs are dedicated co-processors optimized for matrix and vector operations computed at half precision.
TPUs are organized in \emph{pods}, which\footnote{In the TPUv4 architecture.} can connect a total of 4096 of TPU chips with 32 GiB memory each, which together makes up to 128 TiB of distributed memory available for use.
TPUs chips inside a pod are connected with dedicated high-speed, low-latency interconnects organized in a 3D torus topology.

\subsection{Common ML Distribution Strategies}\label{sec:distribution_strategies}

Various methods for distributing Machine Learning workloads have been discussed in the literature~\cite{alqahtani2019distributedml} and most Machine Learning (ML) frameworks provide consistent APIs implementing multiple distribution schemes through a consistent interface. This section highlights some common distribution paradigms focusing on the techniques used to scale DeepWalk using commodity hardware (which we refer to as HUGE-CPU) and TPUs (HUGE-TPU). 

TensorFlow provides the \href{https://www.tensorflow.org/guide/distributed_training}{tf.distribute.Strategy} abstractions to enable users to separate model creation from the training runtime environment with minimal code changes. Two common strategies are the \href{https://www.tensorflow.org/guide/distributed_training#parameterserverstrategy}{Parameter-Server} (PS) strategy and \href{https://www.tensorflow.org/guide/distributed_training#multiworkermirroredstrategy}{Multi-Worker Mirrored Strategy}.

\subsubsection{Parameter-Server Strategy}\label{sec:ps_strategy}
In the context of graph embedding, using a PS strategy is useful for representing a large embedding table. The PS strategy defines two compute pools of potentially heterogeneous hardware that the user can access. One pool contains machines labeled "parameter servers" and the other pool's machines are named "workers". A model's trainable variables are sharded across the machines in the parameter-server pool which serve requests, potentially over a network, both for the values of these variables and to update them. For graph embedding, machines in the worker pool asynchronously receive batches of examples, fetch the necessary embedding rows from parameter servers over a network, compute gradients and push updates back to parameter server machines.

\subsubsection{Multi-Worker Mirrored Strategy}\label{sec:multi_mirrored}
The \href{https://www.tensorflow.org/guide/distributed_training#multiworkermirroredstrategy}{Multi-Worker Mirrored Strategy} replicates all variables in the model on each device in a user defined pool of worker machines. A (potentially) large batch of input examples is divided among the multiple workers and proceed to compute gradients using their smaller per-replica batches. At the completion of a single step, gradients across the replicas are aggregated and all variable copies are updated synchronously. While this can accelerate computationally heavy workloads, compared to the parameter server architecture, this design has limited use in the context of Graph Embedding. Replicating embedding tables across multiple machines introduces unwanted redundancy and memory consumption.

\subsubsection{TPUStrategy and Accelerated TPU Embedding Tables}\label{sec:tpu_strategy}
Training a model (or graph embedding) in TensorFlow using TPU hardware, the \href{https://www.tensorflow.org/guide/distributed_training#tpustrategy}{TPUStrategy} is very similar to the \href{https://www.tensorflow.org/guide/distributed_training#multiworkermirroredstrategy}{MultiWorkerMirroredStrategy}. A user defines a desired TPU topology, a slice of a POD that can be thought of as a subset of interconnected processing units. Under the TPUStrategy, trainable variables are copied to all TPU replicas and large batches of examples are divided into smaller per-replica batches and distributed to available replicas and gradients are aggregated before a syncronous update. Normally, this distribution paradigm would limit the scalability of models that define large embedding tables. However, TPUs are capable of sharding embedding layers over all devices in an allocated topology and leverage high bandwidth interconnections between replicas to support accelerated sparse look-ups and gradient updates. Accelerated embedding tables are exposed in TensorFlow using the \href{https://www.tensorflow.org/api_docs/python/tf/tpu/experimental/embedding/TPUEmbedding}{tf.tpu.experimental.embedding.TPUEmbedding} (TPUEmbedding) layer and are the primary mechanism for scaling DeepWalk training on TPUs. 

\section{Method}\label{sec:method}

We scale the DeepWalk algorithm to embed extremely large-scale graphs using two methods. The first, called \dwcpu{}, uses only commodity hardware whereas the second, \dwtpu{}, leverages modern TPUs for increased bandwidth and performance gains. Figure~\ref{fig:cpu_system} visualizes the parameter-server architecture of \dwcpu{}. The details of parameter-server architecture are covered in section~\ref{sec:dwcpu_design}. Figure~\ref{fig:tpu_system} illustrates the TPU system design behind \dwtpu{} and is detailed in section~\ref{sec:tpu_design}.

\subsection{Preprocessing}
One key observation is that most positional graph embedding systems cannot generate useful embeddings for nodes with less than two edges.
Specifically, nodes with no edges are generally not well defined by embedding algorithms, and similarly, positional embeddings of nodes with only one edge are totally determined by the embedding of their single neighbor.
Therefore, we typically prune the input graph, eliminating nodes with degree less than two. 
In our experiments, we only prune once though the pruning operation itself may introduce nodes that fall below the degree threshold.

\subsection{Sampling}\label{sec:sampling}
After preprocessing the graph, we run random walk sampling to generate co-occurrence tuples that will be used as the input to the graph embedding system. 

A high-level overview of the distributed random walk sampling is provided in Algorithm~\ref{alg:sampling}. The input to the sampling component is the preprocessed graph and the output are TensorFlow Examples containing co-occurrence tuples extracted from the random walks. The implementation of the distributed random walk sampling algorithm is implemented using the distributed programming platform FlumeC++~\cite{flumejava}. 

In the initialization phase, the distributed sampler takes as input the $\mathcal{N}$ nodes of the graph and replicates them $\gamma$ times each to create the seeds of~$\gamma|\mathcal{N}|$ walks it will generate (Line~1).
Next, the random sampling process proceeds in an iterative fashion, performing $k$ joins which successively grow the length of each random walk (Lines~2-4).
Each join combines the walk with the node at its end point.
\footnote{This join is necessary, as the system must support graphs which are too large to fit in the memory of a single machine.}
After joining the end of the walk with its corresponding node from the graph $G$, sampling of the next node occurs (Line~4).
We note that many kinds of sampling can be used here to select the next node at this step -- including uniform sampling, random walks with backtracking, and other forms of weighted sampling.
For the results in this paper, we consider the case of using uniform sampling.
A final GroupBy operation is used to collapse the random walks down to co-occurrence counts between pairs of nodes as a function of visitation distance (Line~7).

The output of the sampling pre-processing step is a sharded series of files encoding a triple: (\sourceid, \destid, \cocounts). 
\sourceid is the node ID of a starting point in the random walk, the \destid is a node ID that was arrived at during the $\gamma$ random walks and \cocounts is a histogram of length \walklen containing the number of times the \sourceid encountered \\ \destid (indexed by the random walk distance of the co-occurrence).

The DeepWalk model defines a graph reconstruction loss that has a ``positive'' and ``negative'' component. The ``positive'' examples are random walk paths that exist in the original graph. ``Negative'' examples are paths that do not exist in the original graph. If desired, the sampling step can be used to generate different varieties of negative samples (through an additional distributed sampling algorithm focusing on edges which \emph{do not} exist).
However, in practice, we frequently prefer to perform approximate random negative sampling ``on-the-fly'' while training.

\begin{algorithm}[t]
\caption{Distributed Random Walk Sampling}\label{alg:sampling}
\algsetup{
linenosize=\small,
linenodelimiter=.
}
\begin{algorithmic}[1]
\REQUIRE A graph $G=(V,E)$, a set of nodes to sample from $\mathcal{N}$, $\gamma$, the number of walks from each node, and $k$ the number of random walks to sample per node.
\ENSURE $\mathcal{C}$, the co-occurrence counts between pairs of nodes in $G$ observed in the walk.
\item[]
\COMMENT{Initialize $k$ walks with each seed}
\STATE $walks \gets \mathcal{N}.\text{repeat}(\gamma)$
\FOR{step = $1 ... k$}
    \item[]
    \COMMENT{Join each walk with the node at its end}
    \STATE $(walks, V) \gets$ $walks$.Join($G$)
    \item[]
    \COMMENT{Extend walk/node pair with new node}
    \STATE $walks \gets (walks, V)$.Sample()
\ENDFOR{}
\item[]
\COMMENT{Group walks from the same seeds together}
\STATE $node\_walks \gets walks$.GroupByKey()
\item[]
\COMMENT{Aggregate co-occurrences for each node's walks}
\RETURN $\mathcal{C} \gets node\_walks$.CombineValues()
\end{algorithmic}
\end{algorithm}

\subsection{Distributed training}

\subsubsection{HUGE-CPU}\label{sec:dwcpu_design}

\begin{figure}
    \centering
    \includegraphics[width=\columnwidth]{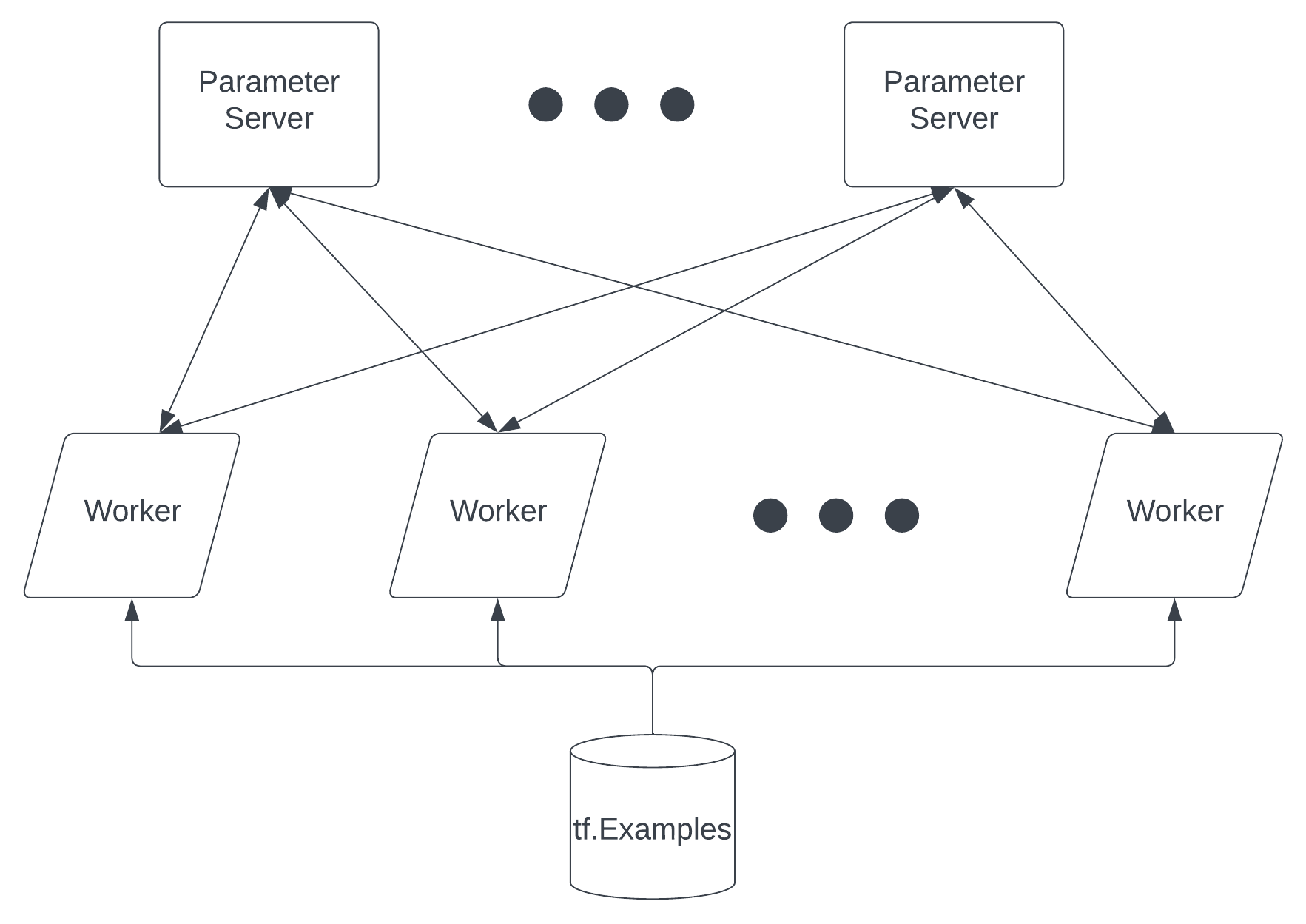}
\caption{System diagram for the Parameter-Server (CPU) based DeepWalk model (HUGE-CPU). Two pools of machines are defined, parameter-servers and workers. Workers asynchronously fetch batches of training examples from disk and collect relevant embedding activations from parameters servers that serve requests for the sharded embedding table. Gradients are computed and updated asynchronously.} 
    \label{fig:cpu_system}
\end{figure}

Figure~\ref{fig:cpu_system} outlines the system design for the HUGE-CPU baseline system architecture. This system leverages distributed training with commodity hardware. Two pools of machines are defined as described in~\ref{sec:ps_strategy}, a cluster of parameter-servers and a pool of workers. During initialization, trainable variables such as the large embedding table are sharded across the machines in the parameter-server pool. Workers distribute and consume batches of training examples from the output of the graph sampling pre-processing step, asynchronously fetch embedding activations from parameter servers, compute a forward pass and gradients and asynchronously push gradient updates to the relevant activations back to the parameter servers. There is no locking or imposed order of activation look-ups or updates. This enables maximum throughput of the system but comes at the cost of potentially conflicting gradient updates.

\subsubsection{HUGE-TPU}\label{sec:tpu_design}

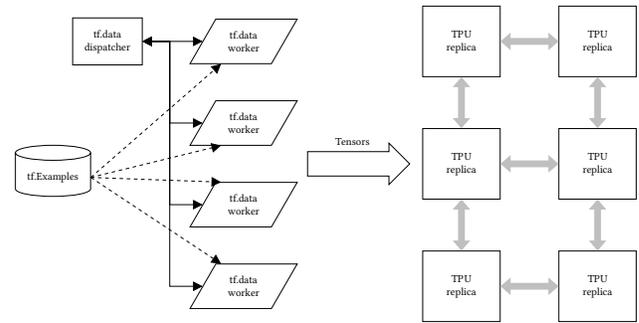
\begin{figure}
    \centering
\resizebox{\columnwidth}{!}{\tikzset{dataworker/.style={draw,trapezium,trapezium left angle = 60,trapezium right angle = 120}}
\tikzset{tpu/.style={draw=black,rectangle, inner sep=20pt}}
\tikzset{tpuarw/.style={>={Triangle[length=3mm,width=5mm]},line width=2mm,draw=gray!50}}
\tikzset{dwarw/.style={{Triangle[width=8pt,length=8pt]}-{Triangle[width=8pt,length=8pt]}, thick}}
\tikzset{twedwarw/.style={dashed, thick, -{Triangle[width=8pt,length=8pt]}}}
\begin{tikzpicture}[every text node part/.style={align=center}, inner sep=10pt, scale=0.8]
\node[draw,cylinder,shape border rotate=90,aspect=0.2] (tfex) at (0,0) {tf.Examples};
\node[draw=black,rectangle] (dispatcher) at (2, 5) {tf.data\\dispatcher};

\node[dataworker] (worker1) at (7,5) {tf.data\\worker};
\node[dataworker] (worker2) at (7,2) {tf.data\\worker};
\node[dataworker] (worker3) at (7,-1) {tf.data\\worker};
\node[dataworker] (worker4) at (7,-4) {tf.data\\worker};

\node[tpu] (r11) at (15, 5) {TPU\\replica};
\node[tpu] (r12) at (15, 0.5) {TPU\\replica};
\node[tpu] (r13) at (15, -4) {TPU\\replica};
\node[tpu] (r21) at (20, 5) {TPU\\replica};
\node[tpu] (r22) at (20, 0.5) {TPU\\replica};
\node[tpu] (r23) at (20, -4) {TPU\\replica};

\path (11,0.5) node [single arrow,draw,minimum height=30mm,label={[yshift=-5] Tensors}]{};

\draw[tpuarw,<->]  (r11.east) -- (r21.west);
\draw[tpuarw,<->]  (r11.south) -- (r12.north);
\draw[tpuarw,<->]  (r21.south) -- (r22.north);
\draw[tpuarw,<->]  (r12.east) -- (r22.west);
\draw[tpuarw,<->]  (r12.south) -- (r13.north);
\draw[tpuarw,<->]  (r22.south) -- (r23.north);
\draw[tpuarw,<->]  (r13.east) -- (r23.west);

\draw[dwarw] (dispatcher.east) -- (worker1.west);
\draw[dwarw] (dispatcher.east) -- +(1, 0) |- (worker2.west);
\draw[dwarw] (dispatcher.east) -- +(1, 0) |- (worker3.west);
\draw[dwarw] (dispatcher.east) -- +(1, 0) |- (worker4.west);

\draw[twedwarw] (tfex.east) -- (worker1.south west);
\draw[twedwarw]  (tfex.east) -- (worker2.south west);
\draw[twedwarw]  (tfex.east) -- (worker3.north west);
\draw[twedwarw]  (tfex.east) -- (worker4.north west);
\end{tikzpicture} }
    \caption{System diagram for accelerated HUGE unsupervised graph embedding. A large embedding table is efficiently sharded over the TPU HBM using TensorFlow TPUEmbedding layer. A cluster of machines that read, parse and randomly sample the input data is leveraged to avoid an input bottleneck. This diagram is illustrative and does not represent the true connectivity of the TPU topology.} 
    \label{fig:tpu_system}
\end{figure}

Figure~\ref{fig:tpu_system} visualizes the system design of distributed training of the DeepWalk embedding model using TPUs after the sampling procedure is complete. The replication strategy used for TPUs in conjunction with their high FLOPS per second requires generating extremely large batches of training examples for every step. The bottleneck in this system is rarely the embedding lookup or model tuning but the input pipeline to generate the large batch size required at every step.

File shards of the sampling data are distributed over the workers in a cluster dedicated to generating input data. The workers independently deserialize the co-occurrence input data and augment the source\_id and destination\_id pairs with negative samples, replicating source\_id and randomly sampling additional destination\_id node IDs uniformly from the embedding vocabulary.

The input cluster then streams the resulting training Tensors to the TPU system which de-duplicates and gathers the relevant embedding activations for the batch and distributes the computational work of computing the forward pass and gradient to the TPU replicas which are then aggregated and used to update the embedding table.

 \section{Experiments}\label{sec:experiments}

\subsection{Experimental Details}

\begin{table}[t]
\centering
\caption{Parameters used for all \dwtpu{} and \dwcpu{} experiments. LWSGD is Stochastic Gradient Descent with a Linear Warmpup and decay learning rate schedule. The schedule is parameterized by four numbers, the number of warmup steps, the final value after warmup, the number of decay steps and the final value after the decay phase at which point the learning rate is held constant.}\label{tbl:dw_params}
\newcolumntype{R}{>{\raggedleft\arraybackslash}X}
\newcolumntype{C}{>{\centering\arraybackslash}X}
\newcolumntype{L}{>{\hsize=.4\hsize}C}
\begin{tabularx}{\linewidth}{@{}CLLC@{}}
\toprule
Parameter & \dwtpu{} & \dwcpu{} \\
\midrule
 \nwalks  & 128  & 128 \\
 \walklen & 3 &  3 \\
 Per-Replica Batch Size & 4096 & 1024\\
 \nnpp & 31 & 3 \\ 
 Global Batch Size & $2^{24}$ & $2^{19}$ \\ 
 LWSGD & (5K, 0.01, 100K, 0.001) & N/A \\
 SGD & N/A & 0.001 \\
\bottomrule
\end{tabularx}
\end{table} 
\subsubsection{Datasets}
For testing the scalability of our methods, we resort to random graphs.
We resort to the standard (degree-free) Stochastic Block Model~\cite{nowicki2001estimation}, which is a generative graph model that divides $n$ vertices into $k$ classes, and then places edges between two vertices $v_i$ and $v_j$ with probability $p_{ij}$ determined from the class assignments.
Specifically, each vertex $v_i$ is given a class $y_i\in\{1,\ldots, k\}$, and an edge $\{v_i, v_j\}$ is added to the edge set $E$ with probability $P_{y_iy_j}$, where $P$ is a symmetric $k\times k$ matrix containing the between/within-community edge probabilities.
Assortative clustering structure in a graph can be induced using the SBM by setting the on-diagonal probabilities of $P$ higher than the off-diagonal probabilities.
For benchmarking, we set $P_{y_iy_j}=q$ iff $i=j$ and to $p$ otherwise.

Complementing our analysis on synthetic benchmark datasets, we also study the performance of the methods on two large real-world graphs: Friendster~\cite{yang2012defining} and OGBN-Papers100m~\cite{hu2020open}.
We report the dataset statistics in Table~\ref{tbl:datasets}.

\subsubsection{Baselines}
First we compare \dwcpu{} and \dwtpu{} with other state-of-the-art scalable graph embedding algorithms: \ie{}~\cite{postuavaru2020instantembedding}, \bg{}~\cite{pbg} and \lne{}~\cite{qiu2021lightne} on an end-to-end node classification task using the \papers{} dataset. We further explore the embedding space quality of each algorithm using both the \papers{} and \friendster{} datasets. Finally we compare embedding space quality metrics as a function of training time to explore the speedups of \dwtpu{} compared to \dwcpu{} using a randomly generated graphs with 100M (\sbmhm{}) and 1B nodes \sbmb{}.

\begin{table}[t]
\centering
\caption{Datasets we use for our experimental studies. We report the total number of nodes and edges in all graphs.}\label{tbl:datasets}
\newcolumntype{R}{>{\raggedright\arraybackslash}X}
\newcolumntype{C}{>{\centering\arraybackslash}X}
\newcolumntype{L}{>{\hsize=.4\hsize}C}
\begin{tabularx}{\linewidth}{@{}RLL@{}}
\toprule
Name & $\lvert{}V\rvert$ & $\lvert{}E\rvert$ \\
\midrule
Friendster & 65.6M & 3612M \\
OGB-Papers100M & 111M & 1616M \\
SBM-10M & 10M & 100M \\
SBM-100M & 100M & 1000M \\
SBM-1000M & 1000M & 10000M \\
\bottomrule
\end{tabularx}
\end{table}
 
\subsection{Parameters for HUGE methods}
Table~\ref{tbl:dw_params} shows the parameters used by the \dwcpu{} and \dwtpu{} methods. The random walk sampling procedure describe in~\ref{sec:sampling} was executed sampling $\gamma=128$ walks per node with a walk length of $k=3$. The set of samples were shared for all experiments involving \dwcpu{} and \dwtpu{} to minimize the affect of random sampling on the results. \nnpp{} is the number of random negative destinations sampled for every "positive" example drawn from the sampling pre-processing step. The global batch size for \dwtpu{} may be computed as $\replicabatch * (1 + \nnpp)$. A step is not well defined for the \dwcpu{} algorithm since workers asynchronously pull variables and push updates. Due to the increased computational power and high bandwidth interconnections between replicas, \dwtpu{} achieves a much higher throughput and global per-step batch size. Training with extremely large batch sizes can be challenging. We have found that a Stochastic Gradient Descent (SGD) optimizer with a linear warmup and ramp down gives good results. \dwcpu{} was also trained with a SGD optimizer but uses a fixed learning rate.

\subsection{Evaluation Metrics}

For all other graphs besides OGBN-Papers100m, there are no ground-truth labels for node classification.
This problem is not unique to publicly available large graphs---in our practical experience, oftentimes there is need to evaluate and compare different embedding models in an unsupervised fashion.
We propose simple unsupervised metrics to compare the embedding quality of different embeddings of a graph.
For the analysis, we $L_2$-normalize all embeddings.

We also report four self-directed metrics for evaluation we use in our production system to monitor the embedding quality. First, {edge signal-to-noise ratio} (\textbf{edge SNR}) defined as:
$$
\mathrm{SNR} = \frac{\mathbb{E}_{u,v\not\in{}E}\left[d(u, v)\right]}{\mathbb{E}_{u,v\in{}E}\left[d(u, v)\right]},
$$
where we approximate the numerator term by taking a random sub-sample of all non-edges.
In our experiments, we also show the entire distribution of \textbf{edge- and non-edge distances}. The intuition behind these metrics is that the distance between nodes that are connected in the original graph (a ``true'' edge) should be ``closer'' than nodes that are not adjacent in the input graph.
Last, we compute the sampled version of the \textbf{edge recall}~\cite{tsitsulin2018verse}.
We sample 100 nodes, pick $k$ closest nodes in the embedding space forming a set $S$.
Then, sampled recall is:
$$
\mathrm{recall}@k(u)=\frac{\vert{}N(u)\cap{}S\vert}{k}.
$$
Despite small sample size, the recall is stable and, coupled with the edge SNR, it is a useful indicator of the reconstruction performance of different graph embedding methods.

\begin{figure*}
\centering
\begin{tikzpicture}
\begin{groupplot}[group style={
                      group name=myplot,
                      group size= 4 by 1,horizontal sep=1.25cm},
                      height=4.5cm,
                      width=.275\linewidth,
                      title style={at={(0.5,0.95)},anchor=south},
                      every axis x label/.style={at={(axis description cs:0.5,-0.2)},anchor=north},]

\nextgroupplot[
 	title = \textbf{Edge SNR},
    bar width=0.45cm,
    ybar=2pt,
    xtick=\empty,
    xlabel=Method,
    ymin=0,
]
\addplot[black,ybar,fill=cycle2,postaction={
        pattern=crosshatch dots
    }] table[y=DWTPU] {data/papers100m_snr.txt};
\addplot[ybar,fill=cycle3,postaction={
        pattern=crosshatch
    }] table[y=DWCPU] {data/papers100m_snr.txt};
\addplot[ybar,fill=cycle4,postaction={
        pattern=north east lines
    }] table[y=IE] {data/papers100m_snr.txt};
\addplot[ybar,fill=cycle5,postaction={
        pattern=horizontal lines
    }] table[y=BG] {data/papers100m_snr.txt};
\addplot[ybar,fill=cycle6,postaction={
        pattern=north west lines
    }] table[y=LNE] {data/papers100m_snr.txt};

\nextgroupplot[
 	title = \textbf{Edge Distance},
    xmin=0, xmax=100,ymin=0, ymax=2,
    xlabel=Pecrentile,
 	legend columns=6,
	legend style={at={(1.15,1.3)},anchor=south,/tikz/every even column/.append style={column sep=0.3cm}},
    legend entries={HUGE-TPU, HUGE-CPU, InstantEmbedding, PyTorch-BigGraph, LightNE},
]
\addplot[mark=*,only marks,mark size=2pt,color=cycle2] table[x expr=100*\thisrow{Quantiles},y=DWTPU] {data/papers100m_edge_distance.txt};
\addplot[mark=square*,only marks,mark size=2pt,color=cycle3] table[x expr=100*\thisrow{Quantiles},y=DWCPU] {data/papers100m_edge_distance.txt};
\addplot[mark=triangle*,only marks,mark size=2pt,color=cycle4] table[x expr=100*\thisrow{Quantiles},y=IE] {data/papers100m_edge_distance.txt};
\addplot[mark=diamond*,only marks,mark size=2pt,color=cycle5] table[x expr=100*\thisrow{Quantiles},y=BG] {data/papers100m_edge_distance.txt};
\addplot[mark=pentagon*,only marks,mark size=2pt,color=cycle6] table[x expr=100*\thisrow{Quantiles},y=LNE] {data/papers100m_edge_distance.txt};

\nextgroupplot[
 	title = \textbf{Non-Edge Distance},
    xmin=0, xmax=100,ymin=0, ymax=2,
    xlabel=Pecrentile,
]
\addplot[mark=*,only marks,mark size=2pt,color=cycle2] table[x expr=100*\thisrow{Quantiles},y=DWTPU] {data/papers100m_non_distance.txt};
\addplot[mark=square*,only marks,mark size=2pt,color=cycle3] table[x expr=100*\thisrow{Quantiles},y=DWCPU] {data/papers100m_non_distance.txt};
\addplot[mark=triangle*,only marks,mark size=2pt,color=cycle4] table[x expr=100*\thisrow{Quantiles},y=IE] {data/papers100m_non_distance.txt};
\addplot[mark=diamond*,only marks,mark size=2pt,color=cycle5] table[x expr=100*\thisrow{Quantiles},y=BG] {data/papers100m_non_distance.txt};
\addplot[mark=pentagon*,only marks,mark size=2pt,color=cycle6] table[x expr=100*\thisrow{Quantiles},y=LNE] {data/papers100m_non_distance.txt};
\nextgroupplot[
 	title = \textbf{Edge Recall},
    xmin=0, xmax=100, ymin=0, ymax=1,
    xlabel=Pecrentile,
]
\addplot[mark=*,only marks,mark size=2pt,color=cycle2] table[x expr=100*\thisrow{Quantiles},y=DWTPU] {data/papers100m_recall.txt};
\addplot[mark=square*,only marks,mark size=2pt,color=cycle3] table[x expr=100*\thisrow{Quantiles},y=DWCPU] {data/papers100m_recall.txt};
\addplot[mark=triangle*,only marks,mark size=2pt,color=cycle4] table[x expr=100*\thisrow{Quantiles},y=IE] {data/papers100m_recall.txt};
\addplot[mark=diamond*,only marks,mark size=2pt,color=cycle5] table[x expr=100*\thisrow{Quantiles},y=BG] {data/papers100m_recall.txt};
\addplot[mark=pentagon*,only marks,mark size=2pt,color=cycle6] table[x expr=100*\thisrow{Quantiles},y=LNE] {data/papers100m_recall.txt};
\end{groupplot}
\end{tikzpicture}
\caption{\label{fig:papers100m} Unsupervised embedding analysis results for \papers{}. We see that \dwtpu{} has superior edge SNR compared to all baselines.}
\end{figure*}
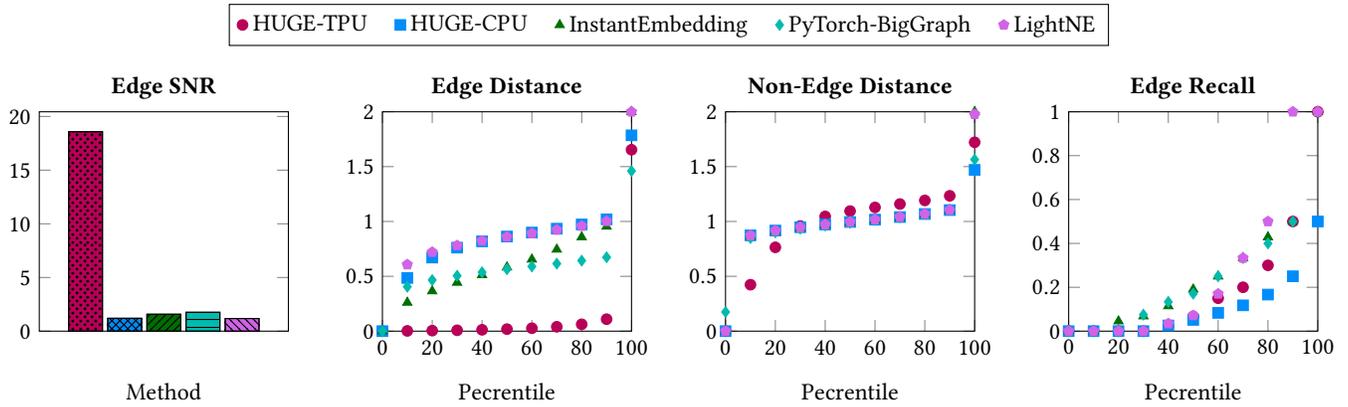 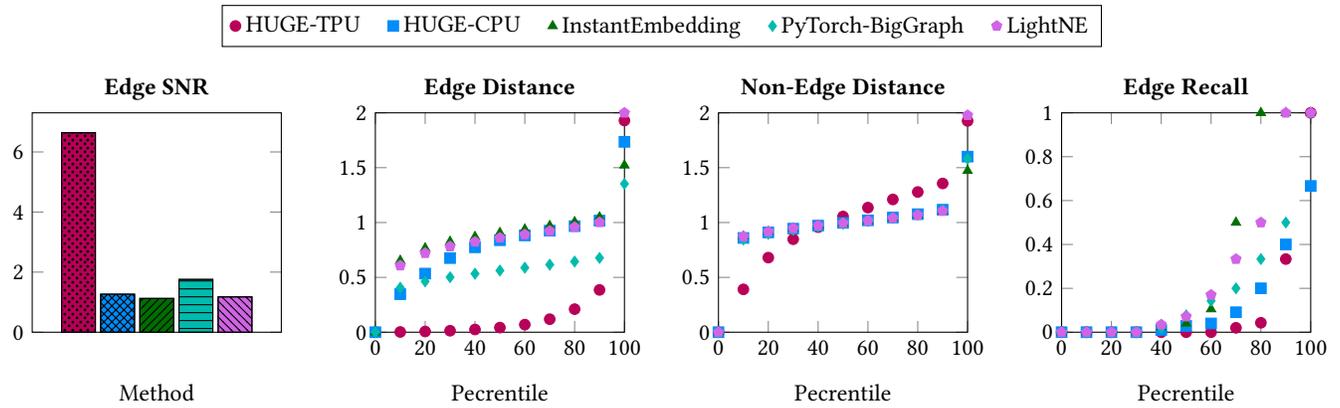
\begin{figure*}
\centering
\begin{tikzpicture}
\begin{groupplot}[group style={
                      group name=myplot,
                      group size= 4 by 1,horizontal sep=1.25cm},
                      height=4.5cm,
                      width=.275\linewidth,
                      title style={at={(0.5,0.95)},anchor=south},
                      every axis x label/.style={at={(axis description cs:0.5,-0.2)},anchor=north},]

\nextgroupplot[
 	title = \textbf{Edge SNR},
    bar width=0.45cm,
    ybar=2pt,
    xtick=\empty,
    xlabel=Method,
    ymin=0,
]
\addplot[black,ybar,fill=cycle2,postaction={
        pattern=crosshatch dots
    }] table[y=DWTPU] {data/friendster_snr.txt};
\addplot[ybar,fill=cycle3,postaction={
        pattern=crosshatch
    }] table[y=DWCPU] {data/friendster_snr.txt};
\addplot[ybar,fill=cycle4,postaction={
        pattern=north east lines
    }] table[y=IE] {data/friendster_snr.txt};
\addplot[ybar,fill=cycle5,postaction={
        pattern=horizontal lines
    }] table[y=BG] {data/friendster_snr.txt};
\addplot[ybar,fill=cycle6,postaction={
        pattern=north west lines
    }] table[y=LNE] {data/friendster_snr.txt};

\nextgroupplot[
 	title = \textbf{Edge Distance},
    xmin=0, xmax=100,ymin=0, ymax=2,
    xlabel=Pecrentile,
 	legend columns=6,
	legend style={at={(1.15,1.3)},anchor=south,/tikz/every even column/.append style={column sep=0.3cm}},
    legend entries={HUGE-TPU, HUGE-CPU, InstantEmbedding, PyTorch-BigGraph, LightNE},
]
\addplot[mark=*,only marks,mark size=2pt,color=cycle2] table[x expr=100*\thisrow{Quantiles},y=DWTPU] {data/friendster_edge_distance.txt};
\addplot[mark=square*,only marks,mark size=2pt,color=cycle3] table[x expr=100*\thisrow{Quantiles},y=DWCPU] {data/friendster_edge_distance.txt};
\addplot[mark=triangle*,only marks,mark size=2pt,color=cycle4] table[x expr=100*\thisrow{Quantiles},y=IE] {data/friendster_edge_distance.txt};
\addplot[mark=diamond*,only marks,mark size=2pt,color=cycle5] table[x expr=100*\thisrow{Quantiles},y=BG] {data/friendster_edge_distance.txt};
\addplot[mark=pentagon*,only marks,mark size=2pt,color=cycle6] table[x expr=100*\thisrow{Quantiles},y=LNE] {data/friendster_edge_distance.txt};

\nextgroupplot[
 	title = \textbf{Non-Edge Distance},
    xmin=0, xmax=100,ymin=0, ymax=2,
    xlabel=Pecrentile,
]
\addplot[mark=*,only marks,mark size=2pt,color=cycle2] table[x expr=100*\thisrow{Quantiles},y=DWTPU] {data/friendster_non_distance.txt};
\addplot[mark=square*,only marks,mark size=2pt,color=cycle3] table[x expr=100*\thisrow{Quantiles},y=DWCPU] {data/friendster_non_distance.txt};
\addplot[mark=triangle*,only marks,mark size=2pt,color=cycle4] table[x expr=100*\thisrow{Quantiles},y=IE] {data/friendster_non_distance.txt};
\addplot[mark=diamond*,only marks,mark size=2pt,color=cycle5] table[x expr=100*\thisrow{Quantiles},y=BG] {data/friendster_non_distance.txt};
\addplot[mark=pentagon*,only marks,mark size=2pt,color=cycle6] table[x expr=100*\thisrow{Quantiles},y=LNE] {data/friendster_non_distance.txt};
\nextgroupplot[
 	title = \textbf{Edge Recall},
    xmin=0, xmax=100, ymin=0, ymax=1,
    xlabel=Pecrentile,
]
\addplot[mark=*,only marks,mark size=2pt,color=cycle2] table[x expr=100*\thisrow{Quantiles},y=DWTPU] {data/friendster_recall.txt};
\addplot[mark=square*,only marks,mark size=2pt,color=cycle3] table[x expr=100*\thisrow{Quantiles},y=DWCPU] {data/friendster_recall.txt};
\addplot[mark=triangle*,only marks,mark size=2pt,color=cycle4] table[x expr=100*\thisrow{Quantiles},y=IE] {data/friendster_recall.txt};
\addplot[mark=diamond*,only marks,mark size=2pt,color=cycle5] table[x expr=100*\thisrow{Quantiles},y=BG] {data/friendster_recall.txt};
\addplot[mark=pentagon*,only marks,mark size=2pt,color=cycle6] table[x expr=100*\thisrow{Quantiles},y=LNE] {data/friendster_recall.txt};
\end{groupplot}
\end{tikzpicture}
\caption{\label{fig:friendster}Embedding analysis results for Friendster. HUGE-TPU achieves the best edge SNR due to better distance distributions.}
\end{figure*} 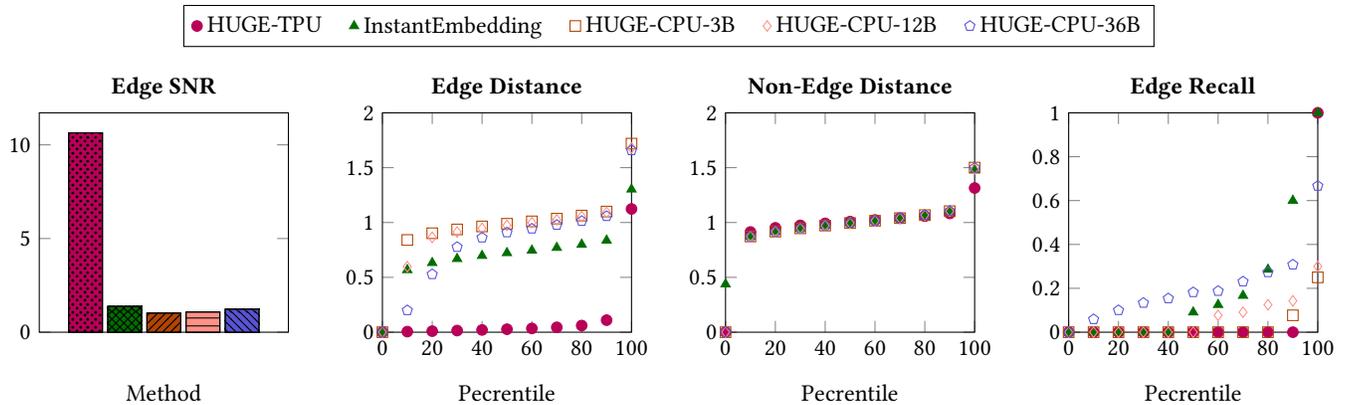
\begin{figure*}
\centering
\begin{tikzpicture}
\begin{groupplot}[group style={
                      group name=myplot,
                      group size= 4 by 1,horizontal sep=1.25cm},
                      height=4.5cm,
                      width=.275\linewidth,
                      title style={at={(0.5,0.95)},anchor=south},
                      every axis x label/.style={at={(axis description cs:0.5,-0.2)},anchor=north},]

\nextgroupplot[
 	title = \textbf{Edge SNR},
    bar width=0.45cm,
    ybar=2pt,
    xtick=\empty,
    xlabel=Method,
    ymin=0,
]
\addplot[ybar,fill=cycle2,postaction={
        pattern=crosshatch dots
    }] table[y=DWTPU] {data/sbm_0100M_snr.txt};
\addplot[ybar,fill=cycle4,postaction={
        pattern=crosshatch
    }] table[y=IE] {data/sbm_0100M_snr.txt};
\addplot[ybar,fill=cycle7,postaction={
        pattern=north east lines
    }] table[y=DWCPU-2h] {data/sbm_0100M_snr.txt};
\addplot[ybar,fill=cycle8,postaction={
        pattern=horizontal lines
    }] table[y=DWCPU-8h] {data/sbm_0100M_snr.txt};
\addplot[ybar,fill=cycle9,postaction={
        pattern=north west lines
    }] table[y=DWCPU-24h] {data/sbm_0100M_snr.txt};

\nextgroupplot[
 	title = \textbf{Edge Distance},
    xmin=0, xmax=100,ymin=0, ymax=2,
    xlabel=Pecrentile,
 	legend columns=6,
	legend style={at={(1.15,1.3)},anchor=south,/tikz/every even column/.append style={column sep=0.3cm}},
    legend entries={HUGE-TPU, InstantEmbedding, HUGE-CPU-3B, HUGE-CPU-12B, HUGE-CPU-36B},
]
\addplot[mark=*,only marks,mark size=2pt,color=cycle2] table[x expr=100*\thisrow{Quantiles},y=DWTPU] {data/sbm_0100M_edge_distance.txt};
\addplot[mark=triangle*,only marks,mark size=2pt,color=cycle4] table[x expr=100*\thisrow{Quantiles},y=IE] {data/sbm_0100M_edge_distance.txt};
\addplot[mark=square,only marks,mark size=2pt,color=cycle7] table[x expr=100*\thisrow{Quantiles},y=DWCPU-2h] {data/sbm_0100M_edge_distance.txt};
\addplot[mark=diamond,only marks,mark size=2pt,color=cycle8] table[x expr=100*\thisrow{Quantiles},y=DWCPU-8h] {data/sbm_0100M_edge_distance.txt};
\addplot[mark=pentagon,only marks,mark size=2pt,color=cycle9] table[x expr=100*\thisrow{Quantiles},y=DWCPU-24h] {data/sbm_0100M_edge_distance.txt};

\nextgroupplot[
 	title = \textbf{Non-Edge Distance},
    xmin=0, xmax=100,ymin=0, ymax=2,
    xlabel=Pecrentile,
]

\addplot[mark=*,only marks,mark size=2pt,color=cycle2] table[x expr=100*\thisrow{Quantiles},y=DWTPU] {data/sbm_0100M_non_distance.txt};
\addplot[mark=triangle*,only marks,mark size=2pt,color=cycle4] table[x expr=100*\thisrow{Quantiles},y=IE] {data/sbm_0100M_non_distance.txt};
\addplot[mark=square,only marks,mark size=2pt,color=cycle7] table[x expr=100*\thisrow{Quantiles},y=DWCPU-2h] {data/sbm_0100M_non_distance.txt};
\addplot[mark=diamond,only marks,mark size=2pt,color=cycle8] table[x expr=100*\thisrow{Quantiles},y=DWCPU-8h] {data/sbm_0100M_non_distance.txt};
\addplot[mark=pentagon,only marks,mark size=2pt,color=cycle9] table[x expr=100*\thisrow{Quantiles},y=DWCPU-24h] {data/sbm_0100M_non_distance.txt}; 
\nextgroupplot[
 	title = \textbf{Edge Recall},
    xmin=0, xmax=100, ymin=0, ymax=1,
    xlabel=Pecrentile,
]
\addplot[mark=*,only marks,mark size=2pt,color=cycle2] table[x expr=100*\thisrow{Quantiles},y=DWTPU] {data/sbm_0100M_recall.txt};
\addplot[mark=triangle*,only marks,mark size=2pt,color=cycle4] table[x expr=100*\thisrow{Quantiles},y=IE] {data/sbm_0100M_recall.txt};
\addplot[mark=square,only marks,mark size=2pt,color=cycle7] table[x expr=100*\thisrow{Quantiles},y=DWCPU-2h] {data/sbm_0100M_recall.txt};
\addplot[mark=diamond,only marks,mark size=2pt,color=cycle8] table[x expr=100*\thisrow{Quantiles},y=DWCPU-8h] {data/sbm_0100M_recall.txt};
\addplot[mark=pentagon,only marks,mark size=2pt,color=cycle9] table[x expr=100*\thisrow{Quantiles},y=DWCPU-24h] {data/sbm_0100M_recall.txt};
\end{groupplot}
\end{tikzpicture}
\caption{\label{fig:sbm-100m}Embedding analysis results for SBM-100M. This figure compares training \dwcpu{} for 3, 12 and 36 billion training examples compared to the results of \dwtpu{} and \ie{}.}
\end{figure*} \begin{figure*}
\centering
\begin{tikzpicture}
\begin{groupplot}[group style={
                      group name=myplot,
                      group size= 4 by 1,horizontal sep=1.25cm},
                      height=4.5cm,
                      width=.275\linewidth,
                      title style={at={(0.5,0.95)},anchor=south},
                      every axis x label/.style={at={(axis description cs:0.5,-0.2)},anchor=north},]

\nextgroupplot[
 	title = \textbf{Edge SNR},
    bar width=0.45cm,
    ybar=2pt,
    xtick=\empty,
    xlabel=Method,
    ymin=0,
]
\addplot[ybar,fill=cycle2,postaction={
        pattern=crosshatch dots
    }] table[y=DWTPU] {data/sbm_1000M_snr.txt};
\addplot[ybar,fill=cycle4,postaction={
        pattern=crosshatch
    }] table[y=IE] {data/sbm_1000M_snr.txt};
\addplot[ybar,fill=cycle7,postaction={
        pattern=north east lines
    }] table[y=DWCPU-2h] {data/sbm_1000M_snr.txt};
\addplot[ybar,fill=cycle8,postaction={
        pattern=horizontal lines
    }] table[y=DWCPU-8h] {data/sbm_1000M_snr.txt};
\addplot[ybar,fill=cycle9,postaction={
        pattern=north west lines
    }] table[y=DWCPU-24h] {data/sbm_1000M_snr.txt};

\nextgroupplot[
 	title = \textbf{Edge Distance},
    xmin=0, xmax=100,ymin=0, ymax=2,
    xlabel=Pecrentile,
 	legend columns=6,
	legend style={at={(1.15,1.3)},anchor=south,/tikz/every even column/.append style={column sep=0.3cm}},
    legend entries={HUGE-TPU, InstantEmbedding, HUGE-CPU-3B, HUGE-CPU-12B, HUGE-CPU-36B},
]
\addplot[mark=*,only marks,mark size=2pt,color=cycle2] table[x expr=100*\thisrow{Quantiles},y=DWTPU] {data/sbm_1000M_edge_distance.txt};
\addplot[mark=triangle*,only marks,mark size=2pt,color=cycle4] table[x expr=100*\thisrow{Quantiles},y=IE] {data/sbm_1000M_edge_distance.txt};
\addplot[mark=square,only marks,mark size=2pt,color=cycle7] table[x expr=100*\thisrow{Quantiles},y=DWCPU-2h] {data/sbm_1000M_edge_distance.txt};
\addplot[mark=diamond,only marks,mark size=2pt,color=cycle8] table[x expr=100*\thisrow{Quantiles},y=DWCPU-8h] {data/sbm_1000M_edge_distance.txt};
\addplot[mark=pentagon,only marks,mark size=2pt,color=cycle9] table[x expr=100*\thisrow{Quantiles},y=DWCPU-24h] {data/sbm_1000M_edge_distance.txt};

\nextgroupplot[
 	title = \textbf{Non-Edge Distance},
    xmin=0, xmax=100,ymin=0, ymax=2,
    xlabel=Pecrentile,
]

\addplot[mark=*,only marks,mark size=2pt,color=cycle2] table[x expr=100*\thisrow{Quantiles},y=DWTPU] {data/sbm_1000M_non_distance.txt};
\addplot[mark=triangle*,only marks,mark size=2pt,color=cycle4] table[x expr=100*\thisrow{Quantiles},y=IE] {data/sbm_1000M_non_distance.txt};
\addplot[mark=square,only marks,mark size=2pt,color=cycle7] table[x expr=100*\thisrow{Quantiles},y=DWCPU-2h] {data/sbm_1000M_non_distance.txt};
\addplot[mark=diamond,only marks,mark size=2pt,color=cycle8] table[x expr=100*\thisrow{Quantiles},y=DWCPU-8h] {data/sbm_1000M_non_distance.txt};
\addplot[mark=pentagon,only marks,mark size=2pt,color=cycle9] table[x expr=100*\thisrow{Quantiles},y=DWCPU-24h] {data/sbm_1000M_non_distance.txt}; 
\nextgroupplot[
 	title = \textbf{Edge Recall},
    xmin=0, xmax=100, ymin=0, ymax=1,
    xlabel=Pecrentile,
]
\addplot[mark=*,only marks,mark size=2pt,color=cycle2] table[x expr=100*\thisrow{Quantiles},y=DWTPU] {data/sbm_1000M_recall.txt};
\addplot[mark=triangle*,only marks,mark size=2pt,color=cycle4] table[x expr=100*\thisrow{Quantiles},y=IE] {data/sbm_1000M_recall.txt};
\addplot[mark=square,only marks,mark size=2pt,color=cycle7] table[x expr=100*\thisrow{Quantiles},y=DWCPU-2h] {data/sbm_1000M_recall.txt};
\addplot[mark=diamond,only marks,mark size=2pt,color=cycle8] table[x expr=100*\thisrow{Quantiles},y=DWCPU-8h] {data/sbm_1000M_recall.txt};
\addplot[mark=pentagon,only marks,mark size=2pt,color=cycle9] table[x expr=100*\thisrow{Quantiles},y=DWCPU-24h] {data/sbm_1000M_recall.txt};
\end{groupplot}
\end{tikzpicture}
\caption{\label{fig:sbm-1000m}Embedding analysis results for SBM-1000M to explore the embedding space quality as a function of training time for \dwcpu{} compared to \dwtpu{} and \ie{}. }
\end{figure*}
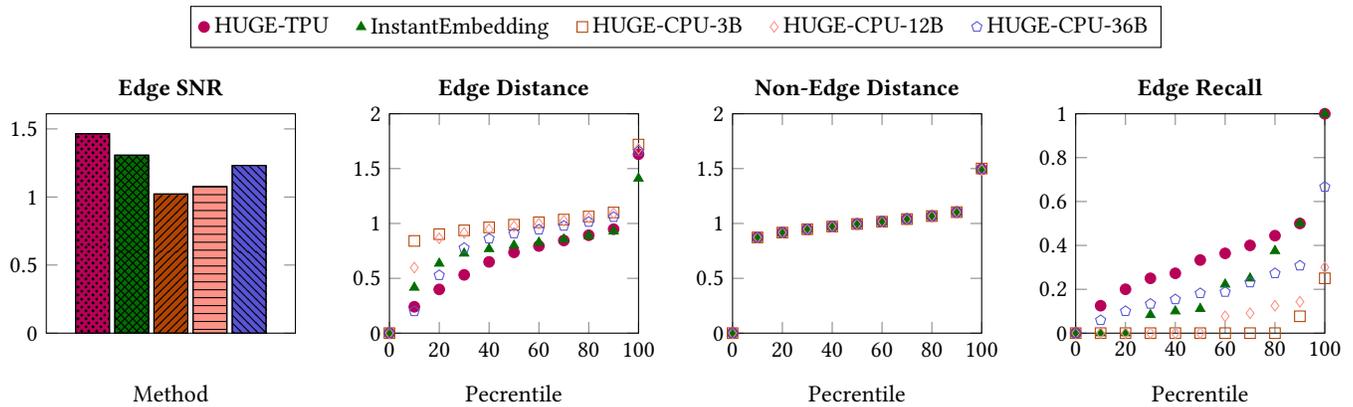 
\subsection{Downstream Embedding Quality}\label{ssec:downstream}

Being the fastest embedding system is not enough -- we also want embedding vectors to be as useful as possible.
Every percentage point of quality on downstream tasks directly translates to missed monetary opportunities.
Therefore, in our experience, when working on scalable versions of algorithms, it is critical to maintain high embedding quality.

\begin{table}[t]
\centering
\caption{Embedding quality as measured by downstream task accuracy, relative speed, and hardware used for four different embedding methods. Speed normalized to the runtime of HUGE-CPU.}\label{tbl:ogb-arxiv}
\newcolumntype{R}{>{\raggedleft\arraybackslash}X}
\newcolumntype{C}{>{\centering\arraybackslash}X}
\newcolumntype{L}{>{\hsize=.4\hsize}C}
\begin{tabularx}{\linewidth}{@{}CLLC@{}}
\toprule
Method & Quality & Speedup & Hardware \\
\midrule
PyTorch-BigGraph & 43.64 & 23.0 & 16x A100 GPUs \\
LightNE & 27.90 & 40.8 & 160 vCPUs \\
InstantEmbedding & 53.15 & 3.5 & 64 vCPUs \\ 
HUGE-CPU & \textbf{56.03} & 1 & 5120 vCPUs \\ HUGE-TPU & \textbf{56.13} & 9.9 & 4x4x4 v4 TPUs \\
\bottomrule
\end{tabularx}
\end{table}
 
To that end, we provide one of the first studies of embedding scalability and performance \emph{across} different hardware architectures.
We compare with the fastest single-machine CPU embedding available~\cite{qiu2021lightne} and industrial-grade GPU embedding system~\cite{pbg}.
Note that these systems have a much more restrictive limit for a maximum number of nodes that they can process.
PyTorch-BigGraph can not process graphs with more than $1.4\times{}10^9$ nodes on the current hardware, assuming a system with the latest-generation GPUs and highest available memory.
LightNE does not have such restriction, but it keeps both the graph and the embedding table in memory.
Because of that, scalability into multi-billion node embedding territory is still an open question for that system.

Table~\ref{tbl:ogb-arxiv} presents the embedding quality results on the OGB-Papers100M dataset.
For measuring the embedding quality, we follow the Open Graph Benchmark evaluation protocol~\cite{hu2020open} with a simple logistic regression model.
We skip tuning the model parameters on the validation set, and report the accuracy of predictions on the test set.
We also report the relative speedup over the CPU DeepWalk embedding implementation.
\dwtpu{} is the only one that maintains the end-to-end classification quality provided by DeepWalk and improves runtime efficiency relative \dwcpu{} by an order of magnitude.

\subsection{Self-directed Embedding Space Evaluation}

To better understand the differences in downstream model performance, we present our self-directed metric for datasets with no ground-truth labels. We analyze the embedding space quality with the proposed metrics comparing \dwcpu{}, \dwtpu{}, \ie{}, \bg{} and \lne{}. 
To that end, we report the results on 2 real and 2 synthetic datasets, presented in Figures~\ref{fig:papers100m}-\ref{fig:friendster} and \ref{fig:sbm-100m}-\ref{fig:sbm-1000m}, respectively.

Interestingly, the results are fairly consistent across all datasets considered.  We see that \dwtpu{} provides superior separation between the distributions of edges and non-edges, achieving a very high edge signal to noise ratio.  We also see that the sampled edge recall metric on is generally much harder to optimize for on very large graphs, and that \dwtpu{} meets or exceeds the performance of its comparable baseline \dwcpu{}.

\subsection{Visualization}

\begin{figure*}
    \centering
    {
    \includegraphics[width=.495\linewidth]{images/viz_all_small.png}
    \hfill
    \includegraphics[width=.495\linewidth]{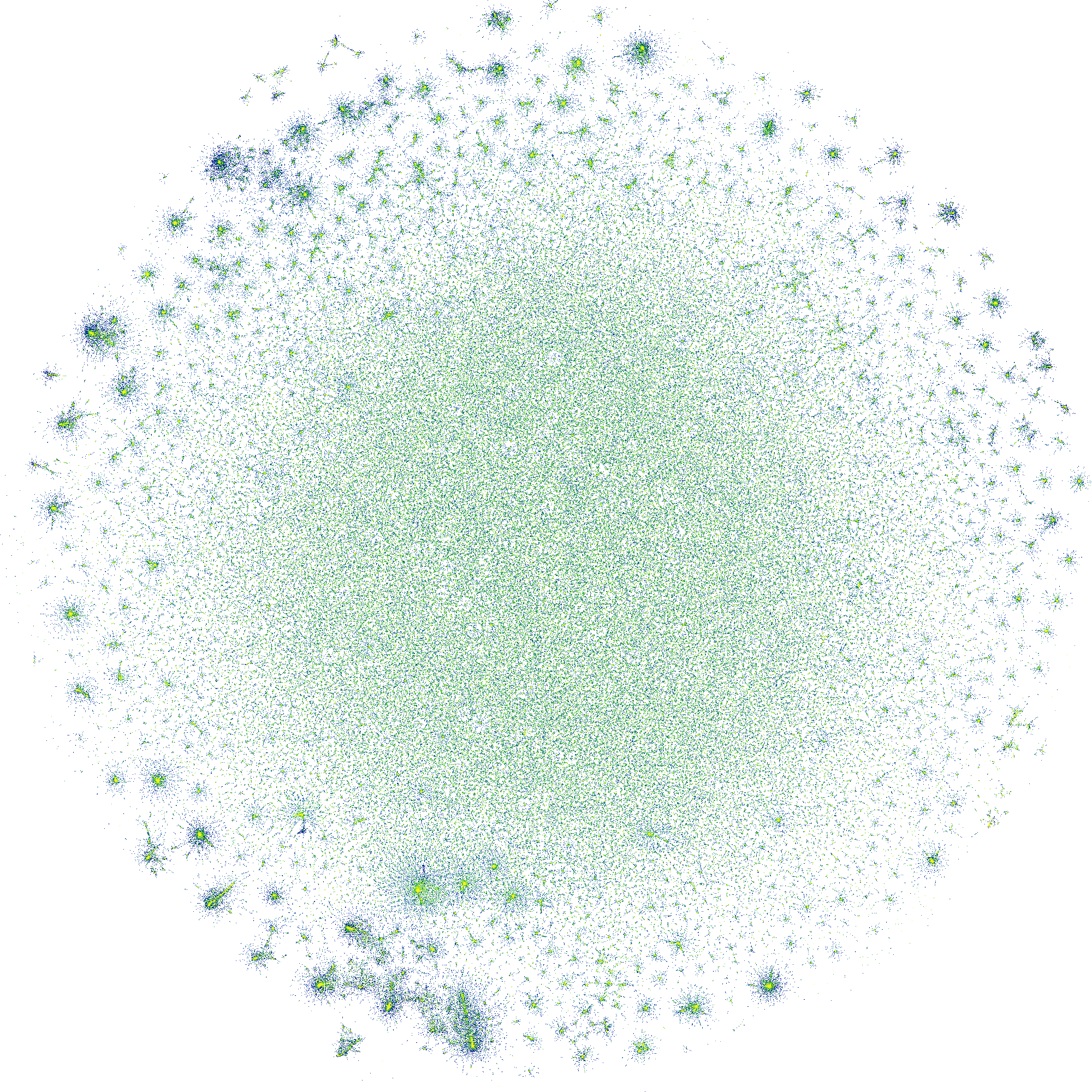}
    }
    \caption{Visualization of the entire embedding space of (left) \dwtpu and (right) LightNE embeddings of the Papers100M dataset, projected via t-SNE with the same parameters. Colors indicate point density. We can observe much better clustered structure in \dwtpu embeddings which directly translates to significantly better downstream prediction quality.\label{fig:viz-large}}
\end{figure*}

In order to better understand our embeddings, we frequently resort to visualizations.
Figure~\ref{fig:viz-large} shows a plot of the \emph{entire} embedding space of OGBN-Papers100M dataset for \dwtpu and LightNE, projected via t-SNE~\cite{van2008visualizing}.
Compared to \dwtpu, the LightNE embedding demonstrated surprisingly poor global clustering structure, which explains its subpar downstream task performance we covered in Section~\ref{ssec:downstream}.

\subsection{Discussion}

\begin{table}[t]
\centering
\caption{The average examples per seconds processed by \dwcpu{} and \dwtpu{} for all reported experiments. \dwcpu{} used 128 Parameter Servers and 128 Workers with 20 cores each. \dwtpu{} was configured with a v4 in a 64 chip configuration. We report the total number of examples processed per second relative to \dwcpu{}}\label{tbl:dw_eps}
\newcolumntype{R}{>{\raggedleft\arraybackslash}X}
\newcolumntype{C}{>{\centering\arraybackslash}X}
\newcolumntype{L}{>{\hsize=.4\hsize}C}
\begin{tabularx}{\linewidth}{@{}p{2cm}C@{}}
\toprule
Method & Relative Examples Per Second \\
\midrule
 \dwcpu{} & 1 \\
 \dwtpu{} &  173x \\
\bottomrule
\end{tabularx}
\end{table} 
While both \dwcpu{} and \dwtpu{} can horizontally scale according to the user configuration, we use the same topologies throughout all experiments. \dwcpu{} uses 128 Parameter Server machines and 128 Workers with 20 cores each and 2GiB of RAM. \dwtpu{} uses a v4 TPU with 64 replicas. The total training examples processed by \dwtpu{} relative to \dwcpu{} for this configuration is shown in table~\ref{tbl:dw_eps}. Since the throughput of \dwcpu{} and \dwtpu{} is fixed for a given topology and batch size, the throughput is constant thought all experiments. 

As shown in table~\ref{tbl:ogb-arxiv}, \dwtpu{} achieves the highest accuracy in the end-to-end node classification task using the \papers{} dataset though \dwcpu{} not far behind. However, while \dwcpu{} is able to scale horizontally to handle extremely large embedding spaces, in-memory and hardware accelerated achieve orders of magnitude speedups compared to \dwcpu{}.

When analyzing the embedding space quality metrics however, in terms of \dwtpu{} consistently achieves superior performance. The distribution of distances between adjacent nodes for \dwtpu{} is typically much smaller than the other methods as is reflected by an \snr{} that is consitently orders of magnitude larger than other methods. The consistently high \snr{} is probably due to the extremely high throughput compared with \dwcpu{}.

To further explore the affect of throughput on the system, we ran \dwcpu{} for multiple fixed number of training steps: 3B, 12B and 36B, while fixing the TPU training time on the \sbmhm{} and \sbmb{} datasets. In relative terms, \dwcpu{}-3B took approximately half the time of \dwtpu{}, \dwcpu{}-12B was trained for 2x the time of \dwtpu{} and \dwcpu{}-36B was trained for 6x the allowed time of \dwtpu{}. We also compare these results with \ie{} to contrast the Deepwalk style embeddings with a matrix factorization graph embedding method. Predictably, the results show that the \dwcpu{} will "converge" or at least approach, over time, the performance of \ie{} in terms of edge/non-edge distributions and recall. However, \dwtpu{} consistently outperforms both \ie{} and \dwcpu{} in terms of \snr{} and edge and non-edge distance distributions even when \dwcpu{} is allowed to train for more than 6x more time than \dwtpu{}.

To summarize, we comprehensively demonstrate the quality and performance of \dwtpu over \dwcpu as well as state-of-the-art industrial-grade systems for graph embeddings.
First, we showed that on a largest-scale labelled embedding data, \dwtpu achieves state-of-the-art performance while being order of magnitude faster than comparable CPU-based system.
We then proceed with unsupervised embedding evaluations we use in deployed production systems at Google.
We show how \dwtpu is competitive in embedding quality over both real and synthetic tasks, only improving its performance compared to the baselines as the size of the graphs increases. \section{Conclusion}\label{sec:conclusion}

In this work we have
examined the problem of scalable graph embedding from a new angle: TPU systems with large amounts of shared low-latency high-throughput memory.
We build a system (HUGE) that does not suffer from key performance issues of the previous work, and greatly simplifies the system design.
HUGE is deployed at Google, in a variety of different graph embedding applications.
Our experiments demonstrate the merits of using accelerators for graph embedding.
They show that the HUGE-TPU embedding is competitive in speed with other scalable approaches while delivering embeddings which are more performant. 
In fact, the embeddings learned with HUGE-TPU are of the same quality as running the full embedding algorithm (with no compromises for its speed). 
\bibliographystyle{ACM-Reference-Format}
\balance
\bibliography{bibliography}

\newpage
\appendix
\section{Experiment setup}

The following setups were used for the experiments with PyTorch-BigGraph and LightNE.

\paragraph{PyTorch-BigGraph.}
\begin{itemize}[noitemsep,topsep=0pt,parsep=0pt,partopsep=0pt]
    \item Google Cloud machine: a2-megagpu-16g, 96 vCPUs, 1.33 TB memory. 16x NVIDIA A100 40GB.
    \item Image: Debian 10 based Deep Learning VM for PyTorch CPU/GPU with CUDA 11.3 M98
    \item BigGraph version as of September 2022 (git@2e94f8a).
\item Configuration Friendster:
            \vspace{-0.5em}
        \begin{lstlisting}
dimension=128, comparator="cos", num_epochs=2, batch_size=100000, num_batch_negs=1000, num_uniform_negs=1000, eval_fraction=0.01, num_gpus=16
        \end{lstlisting}
    \item Configuration Papers100m:
        \vspace{-0.5em}
        \begin{lstlisting}
dimension=128, comparator="cos", num_epochs=3, batch_size=100000, num_batch_negs=1000, num_uniform_negs=1000, eval_fraction=0.01, num_gpus=16
        \end{lstlisting}
\end{itemize}

\paragraph{LightNE}
\begin{itemize}[noitemsep,topsep=0pt,parsep=0pt,partopsep=0pt]
    \item Google Cloud machine: m1-ultramem-160, 160 vCPUs, 3.84 TB memory.
    \item Image: Debian 11, v20220920
    \item LightNE version: version as of September 2022 (git@c43afc4)
\item Configuration:
            \vspace{-0.5em}
        \begin{lstlisting}
-walksperedge 10000 -walklen 3 -step_coeff 1,1,1 -rounds 1 -s -m -ne_method netsmf -rank 256 -dim 128 -order 10 -sample_ratio 17 -mem_ratio 0.5 -negative 1 --sparse_project 0 -sample 1 -upper 0 -analyze 1 
        \end{lstlisting}
\end{itemize} 
\end{document}